\documentclass[10pt, a4paper]{article}

\usepackage[final]{lrec2026} 

\usepackage[frozencache,cachedir=.]{minted}

\usepackage{caption}

\usepackage[]{mdframed}

\usepackage{underscore}
\usepackage{titlecaps}
\Addlcwords{the and but or nor for a an at by to in on with of}
\let\oldsubsection\subsection
\renewcommand{\subsection}[1]{\oldsubsection{\titlecap{#1}}}

\usepackage[autostyle]{csquotes}
\usepackage[english]{babel}

\usepackage{enumitem}

\usepackage[skip=0pt, indent]{parskip}

\usepackage{adjustbox}

\usepackage{booktabs}
\usepackage{tablefootnote}
\usepackage{multirow}

\usepackage{circledsteps}

\newcommand{\cri}[1]{{\emph{#1}}}
\newcommand{\redemption}{\cri{redemption at maturity}}
\newcommand{\instrument}{\cri{type of instrument}}
\newcommand{\principal}{\cri{principal amount}}
\newcommand{\simple}{``simple'' criteria}
\newcommand{\complex}{\emph{complex} criteria}

\newcommand{\pr}{prospectus}

\newcommand{\prs}{prospectuses}
\newcommand{\PRS}{Prospectuses}

\newcommand{\bp}{base prospectus}

\newcommand{\bps}{base prospectuses}

\newcommand{\sd}{master data} 

\newcommand{\llama}{Llama-3.3-70B-Instruct}
\newcommand{\cohere}{Command-R 08-2024}
\newcommand{\mistral}{Mistral-Small-3.1-24B-Instruct-2503}

\title{LLM-Based Examination of Eligibility Criteria from Securities Prospectuses at the German Central Bank}

\name{Serhii Hamotskyi, Akash Kumar Gautam, Christian Hänig} 

\address{Anhalt University of Applied Sciences \\
         \{serhii.hamotskyi, akash-kumar.gautam, christian.haenig\}@hs-anhalt.de\\}

\abstract{ 
Verifying the eligibility of securities as collateral is a key responsibility of the German Central Bank.
However, manually verifying these assets against legal and financial criteria within lengthy, semi-structured, and often bilingual prospectuses is a resource-intensive task. 
While previous efforts utilized traditional Named Entity Recognition (NER) for information extraction, these methods can struggle with OCR noise, linguistic variance, and rigid span-based constraints, 
and the need for manually annotated training data for each relevant annotation type.
In this paper, we present the first case study applying Large Language Models (LLMs) to the eligibility examination process, shifting the paradigm toward a generative Information Extraction pipeline. Our approach decomposes the task into extraction, normalization, and interpretation, allowing for greater flexibility in handling noisy text and interleaved German-English content. We further introduce a value-based evaluation methodology using LLM-as-a-judge, which offers a more semantic assessment than location-based metrics. Our results demonstrate that LLM-based systems achieve high precision (up to 91\%) in document-level eligibility, exhibiting a conservative operating profile that minimizes false acceptance.\\
\newline \Keywords{Large Language Models, Information Extraction, LLM-as-a-Judge, Financial NLP.} }

\begin{document}

\maketitleabstract


\section{Introduction}
As the central bank of the Federal Republic of Germany and a core member of the Eurosystem, the \emph{Deutsche Bundesbank} is responsible for implementing monetary policy and providing liquidity to the financial system. These operations are conducted as credit transactions: they must be backed by collateral to protect the central bank from financial risk. 
The acceptance of a security as collateral depends on its \textbf{eligibility}, which is based on specific legal and financial criteria to ensure that only high-quality assets are pledged~\cite{europeancentralbank.eurosystemcollateral2017}.

The eligibility of securities is assessed on the basis of their \textbf{\prs{}}, which can be hundreds of pages long.
Thousands of securities are issued annually, and verifying their eligibility is a time-consuming and tedious process, making eligibility estimation a prime target for automation.

The difficulty lies in the semi-structured nature of the source material, with evidence being scattered across the entire document and expressed through dozens of different conventions. 
\PRS{} can be bilingual, with English or German interleaved or presented in parallel columns, requiring models that are robust to language switching, ideally — able to use the information from both. 

Previous work by~\citet{hanignlpbaseddecision2023} addressed this challenge by developing a Decision Support System 
that models the task as a Named Entity Recognition (NER) problem, solved using Transformer-based models~\cite{vaswani2017attention}, achieving good results on most criteria. 
Nevertheless, that approach introduced several constraints, primarily: it required extensive  manual annotation to provide necessary supervision for all relevant annotation types, and the resulting models were sensitive to the rigid boundaries of text spans (which made them fragile when encountering OCR artifacts or financial language different from its training set).

In this paper, we present the first case study applying Large Language Models (LLMs) to the eligibility examination process at the German Central Bank, shifting the paradigm from traditional token classification to a generative Information Extraction approach. 

The main \textbf{contributions} are as follows.
\begin{itemize}[topsep=0px] 
  \setlength{\itemsep}{0pt}
  \setlength{\parskip}{0pt}
    \item Presenting a multi-stage generative Information Extraction pipeline 
    that allows 
    the handling of 
    linguistic structures that 
    token-classification models may struggle to process
    \item Introducing a value-based evaluation methodology using LLM-as-a-judge, 
    resistant to OCR noise and linguistic variance
\end{itemize}

\noindent
Although specialized domain models have shown good results in financial tasks, our study focuses on the zero-shot and instruction-following capabilities of high-performance general-purpose models: \llama{}~\cite{grattafiori2024llama} and
Cohere \cohere\footnote{\href{https://huggingface.co/CohereLabs/c4ai-command-r-08-2024}{https://hf.co/CohereLabs/c4ai-command-r-08-2024}} for inference, and
Mistral Small 3.1 Instruct\footnote{\href{https://huggingface.co/mistralai/Mistral-Small-3.1-24B-Instruct-2503}{https://hf.co/mistralai/Mistral-Small-3.1-24B-Instruct-2503}} for evaluation.

\section{Examination of Eligibility Criteria}
In the context of this case study, eligibility is determined by 
6
criteria\footnote{In~\citet{hanignlpbaseddecision2023} two more are listed, and were excluded due to having few examples in the training data.}, 
all of which must be fulfilled for the \pr{} to be eligible.
The descriptions that follow are simplifications 
and do not 
fully reflect the Eurosystem eligibility criteria\footnote{\href{https://eur-lex.europa.eu/legal-content/EN/TXT/PDF/?uri=CELEX:32014O0060}{https://eur-lex.europa.eu/legal-content/EN/TXT/PDF/?uri=CELEX:32014O0060}}.

\begin{description}[topsep=5px]
  \setlength{\itemsep}{0pt}
  \setlength{\parskip}{0pt}
    \item[Currency] One of: EUR, USD, GBP, JPY
    \item[Type of instrument] Only certain types of financial instruments are allowed (e.g. stocks are not)
    \item[Principal amount] Only fixed and unconditional amounts 
    
    \item[Redemption (amount) at maturity] The principal amount must be repaid in full at bond maturity
    \item[Coupon] Only certain coupon structures allowed 
    \item[Status] Not subordinated to other debt. 
\end{description}

\noindent
The first 4 criteria will be referred to as \simple, as they depend only a single extracted entity from the document text. The last two \complex{} --- \cri{coupon} and \cri{status} --- are determined using a decision tree using \emph{multiple} types extracted from the \pr, and \sd{} (asset type, issuer group, issuer date).


\bigskip

\noindent
Each security is described by three data points.
\textbf{\PRS{}} are PDF files describing the terms and conditions governing the issuance of the security.
A \textbf{\bp{}}, if present, is considered part of the \pr. (An issuer might have a \bp{} with overall standard terms, and issue \prs{} for each individual security containing information applicable to it specifically.) 
Crucially, the annotations in \bps{} are available during inference and, in most cases, take precedence over data predicted from the \pr. 
Each \pr{} is always accompanied by \textbf{\sd{}} (\emph{Stammdaten}) — additional metadata about each \pr, including its \textbf{eligibility} and the name of its \bp{} if present. These (except eligibility) are also available during inference.

\section{Related Work}

Information Extraction (IE) from financial documents is a rapidly evolving field, recently transitioning from traditional discriminative models to generative Large Language Models (LLMs). 

\paragraph{Information extraction}
\citet{colakoglu-etal-2025-problem} systematically evaluates different building blocks for LLM-based IE in layout-rich documents, including input formats, prompt structures, and cleanup/postprocessing steps applied to LLM output.
\citet{chenevaluationprompt2025} discusses the importance and evaluation of prompt engineering as applied to IE, with a focus on OCR-derived data.

\citet{lu-huo-2025-financial} is a recent \enquote{systematic evaluation of state-of-the-art LLM and prompting methods} applied specifically to financial NER, comparing them to Transformer-based models, and finding that the latter consistently outperform generic LLMs, with prompt design and in-context learning narrowing the gap.

\paragraph{Pre-trained Models for Financial Domain}
Though we focus on zero-shot general-purpose models, domain-specific finetuning has often been used for similar tasks and shown to be able to outperform significantly larger models.

This includes early efforts like FinBert~\cite{yangfinbertpretrained2020},  
German-language financial models~\cite{kozaeva-etal-2024-development}, and more recent multilingual suites like LLM Pro Finance~\cite{caillautllmpro2025}.
See~\citet{leecomprehensivereview2024} for a deeper review of finance-specific, including multimodal, LLMs.

\paragraph{Benchmarking and Evaluation}

FinBen~\cite{xiefinbenholistic2024} is a holistic financial benchmark for LLMs, spanning 24 tasks including information extraction. 
As tasks move beyond simple classification, ``LLM-as-a-judge'' has emerged as a vital tool for semantic evaluation. Current research focuses on improving the reliability and consistency of these automated judges to replace or augment rigid, offset-based metrics~\citep{gusurveyllmasajudge2025}.


\section{Data}

\begin{figure*}[t]
\begin{center}
\includegraphics[width=1.0\textwidth]{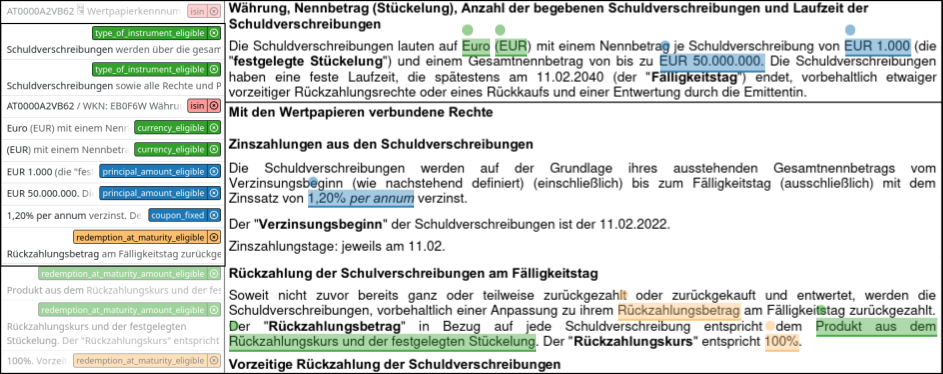}
\caption{Sample annotated paragraphs. Note the varying length and complexity of different types, as well as annotations of the same type present in multiple locations.}
\label{fig:annos}
\end{center}
\end{figure*}

\subsection{The Dataset}
\paragraph{Dataset Summary}
The dataset we use was originally created in the context of \citet{hanignlpbaseddecision2023}, who modeled the task as NER and annotated it as such. 

The dataset is composed of 413 \prs{}, split into a train set 
(\textbf{268} 
\prs) 
and a test set (\textbf{145}).
The test set \prs{} were annotated twice (each\footnote{A small number were unusable for technical reasons.} by two different annotators), resulting in
\textbf{285}
\emph{annotated} documents,
of which 
82 
(\textasciitilde29\%)
were ineligible.

The PDF files are sourced from the FinCorpus-DE10k~\citep{hamotskyifincorpusde10kcorpus2024} corpus and share its characteristics, particularly regarding PDF layouts, OCR artifacts, and the parallel presentation of English-German text.

\paragraph{Bilinguality and PDF Layouts}
All documents are in German, and about a third are bilingual (English and German). Only the German text is annotated and considered primary. 
In the bilingual documents, different layouts are possible, including languages in separate columns or interleaved line by line.
Tables, footnotes and checkboxes are present. 

\subsection{Annotations}
\label{sec:annotations}
There are 18 annotation types. A partial annotated document is shown on Figure~\ref{fig:annos}.

In the case of \simple, the presence or absence of an annotation of a certain type is enough to determine whether the criterion is fulfilled. 
For example, if the document text states that the currency for the security is EUR (one of the eligible currencies), the span was marked with the annotation type \mbox{\texttt{currency_eligible}}; the criterion \cri{currency} is considered fulfilled. 
Crucially, the intent was to annotate the place with the \textit{evidence} rather than mere entity mentions. A document may mention several currencies, but only one defining the security's denomination constitutes evidence for eligibility.
The absence of that annotation in a document implies that either that information is absent, or that the currency is not an eligible one. 

\cri{Currency} here is useful as an example, but most of the other annotation types are more complex, longer (more than 30 words in some cases) and exhibit larger variance. 
A span containing the currency name is easily normalizable into a standard format for further processing (€,~Euro,~...~\textrightarrow~EUR) using a rule-based system, but not all extracted types are.

Not the entire \pr{} was annotated — only enough to make an eligibility estimation. The rest was marked with a special \texttt{Block} annotation, to signal that NER models shouldn't be trained on that text because it might contain unannotated entities. 
An important side effect of this was that different annotators could find evidence for the same criterion in different places of the document.
This had implications for both extraction and evaluation, see~\ref{para:biling}.

\section{Methodology}

\begin{figure*}[t]
\begin{center}
\includegraphics[width=0.75\textwidth]{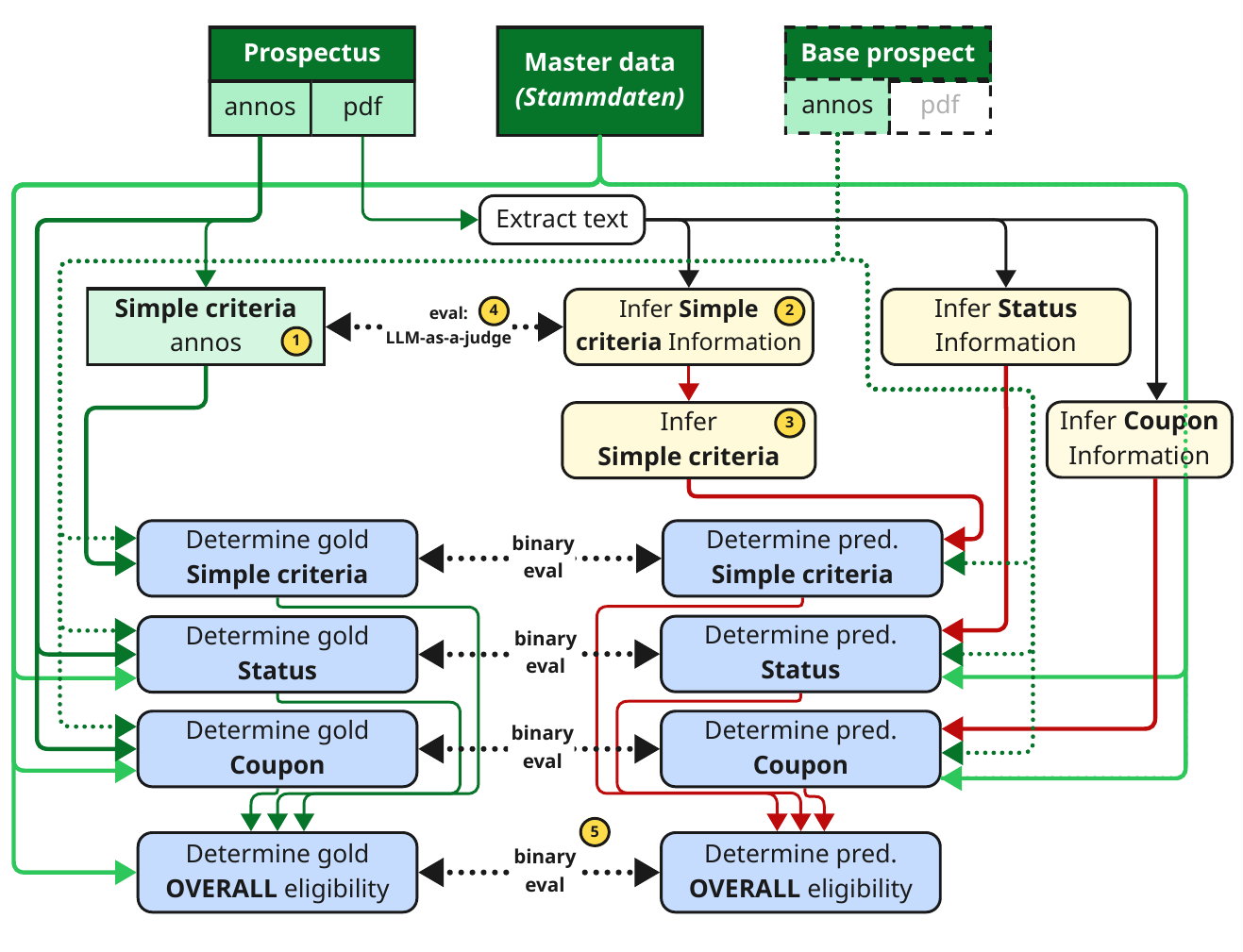}
\caption{Simplified flow of the system. Green arrows represent ground truth data (or data deterministically derived from it), red arrows denote predictions, and dashed lines are \bp{} annotations that may be absent. Green rectangles are ground truth data objects, deterministic processes are in blue, and LLM blocks start with \emph{Infer} and are yellow.}
\label{fig:flow}
\end{center}
\end{figure*}

\subsection{Architecture}
The main building blocks are shown in the diagram on Figure~\ref{fig:flow}.
LLMs are used for \textbf{extraction, normalization}, and \textbf{interpretation}. Their results are then processed algorithmically into the final criteria determination using Python.

\paragraph{Ground truth}
First, the annotations from both the \pr{} and \bp{} (if present) are parsed. The \pr{} annotations are processed into ground truth criteria decisions and are used for evaluation; the \bp{} annotations are available during inference, and are provided to the blocks that determine the final predicted criteria. 

\paragraph{PDF preprocessing}
\label{para:preprocessing}
Before inference, the \pr{} PDF is converted into Markdown using Docling~\cite{livathinosdoclingefficient2025}. 

The dataset already had extracted text created during the annotation process.
That text 
contained many atypical or private-use-area Unicode code points (often in connection with checkboxes) and large amounts of inconsistent spacing.
This led to unpredictable behavior during inference, including repetitions and unreliable JSON generation, especially by the \cohere{} model on longer \prs. 
Normalizing the text fixed these issues, and we quickly found that Docling offers good quality extracted text that requires no additional processing from our side. The markdown conversion preserves formatting, providing additional information about the structure of the document.

The offsets of the annotations referred to the previous (original) extracted text, and were thus rendered invalid. This had implications for evaluation but not for inference, since text extraction would have been performed anyway when classifying new \prs.

\paragraph{Inference}
Data is first \textbf{extracted} and \textbf{normalized} using an LLM.

For example, for the \simple, a structure similar to Listing~\ref{lst:simplecriteriainfo} is first extracted 
(\Circled{2} on the diagram). 
For each criterion, it contains the value (required information in a normalized form), the raw value (exactly as stated in the text), and a quote from the source document with the surrounding context. 

\begin{listing}
\begin{minted}[fontsize=\footnotesize]{json}
"principal_amount": {
  "raw_value": "up to 10.000,00€",
  "value": "up to EUR 10.000",
  "evidence": {
    "source": "Prospectus",
    "exact_quote": "in an aggregate principal 
    amount of up to 10.000,00€ divided 
    into up to 1,500 Pfandbriefe"
  },
},
"currency": {...},
"redemption_at_maturity": {...},
"type_of_instrument": {...}
\end{minted}
\caption{Extracted (\texttt{raw_value}) and normalized (\texttt{value}) for \principal.}
\label{lst:simplecriteriainfo}
\end{listing}

For the \simple, a second inference step (\Circled{3}) \textbf{interprets} the extracted data into criteria predictions, for example checking whether the principal amount inferred in the previous step is fixed (and therefore valid), with results similar to Listing~\ref{lst:simplecriteria}. 

\begin{listing}
\begin{minted}[fontsize=\footnotesize]{json}
"principal_amount": {
  "eligible": true,
  "reason": "The principal amount is 
      stated as 'up to EUR 10.000', 
      which only caps the issuance 
      volume and does not indicate 
      variability.",
  "details": {"source": {...}}
},
"currency": {...},
"redemption_at_maturity": {...},
"type_of_instrument": {...}
\end{minted}
\caption{Interpretation of extracted data into an eligibility prediction.}
\label{lst:simplecriteria}
\end{listing}

For the ``complex'' criteria, the information required is extracted and interpreted in a single step.

\paragraph{Processing of the results}
Finally, a final decision on each criterion is made based on the inferred result, \bp{} annotations if present, and \sd{} (for the ``complex'' criteria).  
All these steps are done in Python.
Lastly, the overall \pr{} eligibility is determined: eligible if all criteria are fulfilled, ineligible otherwise.

\subsection{Text extraction with LLMs}
After initial tests, we focused on two models: \llama\footnote{\href{https://huggingface.co/meta-llama/Llama-3.3-70B-Instruct}{https://hf.co/meta-llama/Llama-3.3-70B-Instruct}} and Cohere \cohere\footnote{\href{https://huggingface.co/CohereLabs/c4ai-command-r-08-2024}{https://hf.co/CohereLabs/c4ai-command-r-08-2024}} (32B).  
While \llama{} served as a high-reasoning baseline, \cohere{} was selected for its multilinguality as well as its specialized training in grounded generation and RAG-specific tasks, which we hypothesized would minimize hallucinations when quoting long financial prospectuses.
Both have a large 128k context length, which allowed us to quote entire \prs{} directly in the prompt, without needing any of the methods used for processing documents longer than the model context.

We used LangChain with structured output to force the models into the required JSON schema. 

For longer documents, \cohere{} often ended up stuck outputting tab characters or incorrectly escaping nested quotes of the documents it cited, returning incorrect JSON as a result. 
Applying a frequency penalty of $0.05$ (in addition to text preprocessing discussed in Section~\ref{para:preprocessing}) mitigated this issue.
For both models we used a temperature of $0.1$.

\section{Evaluation}
\label{sec:evaluation}
Evaluation is performed on three levels: 
(i) overall (per-document) \pr{} eligibility (\Circled{5} on Figure~\ref{fig:flow}),
(ii) criteria results, and 
(iii) comparison of the extracted values to the annotations (\Circled{4}).


\subsection{Document-level and criteria evaluation}
\paragraph{Document-level} A \pr{} is eligible if and only if all the criteria are fulfilled. The scores are shown on Table~\ref{tab:overall}. 
\paragraph{Criteria evaluation}
Same as document-level eligibility, this was evaluated as a binary classification task.
The ``complex'' criteria  required different extracted information depending on the
\sd{} values (in some cases none at all if the process was fully deterministic).
As a result, their scores have a less direct dependence on LLM predictions. 
The results are shown on Table~\ref{tab:critprf}.

\subsection{Evaluating the extracted values}
\label{sec:eval-extracted}
Evaluating the individual criteria measures the bottom line, but it is not the complete picture — a criterion can be correct for the wrong reasons. For instance, only a fixed/invariable \principal{} is eligible. Extracting some different amount would set the criterion to the correct value as long as that amount is fixed. 
Thus, the extracted data itself also needs to be evaluated.

The NER classifier in \citet{hanignlpbaseddecision2023} used the standard \textbf{offset-based} evaluation, comparing the locations of the predicted entities to the annotated ones. 

Position-based evaluation is inherently flawed for long, semi-structured documents where the same evidence may appear redundantly (and as noted in~\ref{sec:annotations}, annotating \emph{all} occurrences was not a goal during annotation). 
A \textbf{value-based} approach, which prioritizes semantic truth over positioning, was used. 

\paragraph{Evaluation setting}
For each field we needed to extract, we compared our (single) extracted value to (potentially multiple) annotations of the corresponding type. Either could be missing if the information was not found in the document.


For both we used the threshold-based approach from~\citet{chenevaluationprompt2025}, which we expanded to handle zero or multiple ground truth annotations and missing predictions.

If at least one annotation of the relevant annotation type (${y_{true}}$) had a similarity of $>80\%$ to our extraction ($y_{pred}$), we considered it a match; see Table~\ref{tab:binary} for the other cases. Each case, then, became either a True Positive, False Positive, True Negative, or False Negative (TP, FP, TN, FN on the table). From these Accuracy, Precision, Recall and F1-Score were calculated. 

\begin{table}
\small 
\begin{center}
\begin{tabular}{ c c c c }
\hline
$\mathbf{y_{true}}$ & $y_{pred}$ & $\max(sim(\mathbf{y_{true}}, y_{pred}))$ & res \\
\hline 
+ & + & $\geq 80\%$ & TP \\ 
+ & + &  $<80\%$ & FN \\
+ & - & N/A  & FN \\
- & + &  N/A & FP \\
- & - &  N/A & TN \\
\hline
\end{tabular}

\caption{\label{tab:binary} +/- denotes presence/absence of at least one element; the max similarity is between the $y_{pred}$ and all $\mathbf{y_{true}}$.}
\end{center}
\end{table}

For calculating the similarity, we used two approaches: fuzzy string matching and LLM-as-a-judge. 
The results are shown on Figure~\ref{fig:fuzzy}. 

\paragraph{Fuzzy string matching}
Following~\citet{chenevaluationprompt2025} we used \texttt{fuzzy.token_set_ratio} of the \texttt{fuzzywuzzy}\footnote{\href{https://pypi.org/project/fuzzywuzzy/}{https://pypi.org/project/fuzzywuzzy/}} package as simple similarity metric. It can match strings partially and is robust to changes in token order.

\paragraph{LLM-as-a-judge}
We drafted custom instructions for every field we extracted, containing both specific rules about what is considered equivalent (e.g., different subtypes of the same financial instrument type) and generic ones (\enquote{equivalence is not affected by OCR noise, formatting, language, singular/plural}). 

The judge-LLM returned scores and the reasons for them, which was crucial for explainability and helpful for improving the evaluation instructions. See~\ref{lst:judge} for an example.

\begin{listing}
\begin{minted}[fontsize=\footnotesize, 
    % ignorelexererrors=true, 
    breakaftersymbolpre={},    % Removes the arrow at the end of the line
    breakaftersymbolpost={},   % Removes the arrow at the start of the next line
    % breakafter={},   
    % prebreak={},
    % postbreak={},
]{json}
"y_true": "Inhaber -schuldverschreibung",
"y_pred": "Schuldverschreibungen",
"llm_match": 1.0,
"llm_match_reason": "An  
    'Inhaberschuldverschreibung' is a
    type of 'Schuldverschreibung';  
    formatting differences are 
    irrelevant."
\end{minted}
\caption{A sample LLM-as-a-judge result. Terminology: \textit{Inhaberschuldverschreibung} (bearer bond), \textit{Schuldverschreibung} (debt security/bond).}
\label{lst:judge}
\end{listing}

The model used was \mistral\footnote{\href{https://huggingface.co/mistralai/Mistral-Small-3.1-24B-Instruct-2503}{https://hf.co/mistralai/Mistral-Small-3.1-24B-Instruct-2503}} with its standard settings. 
The evaluation itself was executed using the \texttt{pydantic-evals}\footnote{\href{https://ai.pydantic.dev/evals/}{https://ai.pydantic.dev/evals/}} framework.

\section{Results}



\begin{table}
\begin{center}
\small
\begin{tabular}{l  c c c c }
\hline
 & Acc. & F1 & Pre. & Rec. \\
\hline 
\citet{hanignlpbaseddecision2023} & 0.60 &	0.72 & 	0.70 & 	0.76 \\ 	
\llama & 0.82 &	0.85 & 	0.90 & 	0.80 \\ 	
\cohere & 0.84 &	0.86 & 	0.91 & 	0.82 	\\
\hline
\end{tabular}
\caption{Per-document eligibility scores. \citet{hanignlpbaseddecision2023} scores provided by author.} 
\label{tab:overall} 

\end{center}
\end{table}

\begin{table*}[t]
\newcolumntype{s}{>{\columncolor[HTML]{AAACED}} p{3cm}}
\centering
\small
\begin{adjustbox}{max width=\textwidth}

\begin{tabular}{@{\extracolsep{\fill}} l  r | rrrr | rrrr} 
\toprule
 & \citet{hanignlpbaseddecision2023} & \multicolumn{4}{c|}{Llama3.3-70B-Instruct} & \multicolumn{4}{c}{Command-R 08-2024} \\
& \textbf{Acc.}$^{\ddag}$ & Precision & Recall & F1 & \textbf{Acc.} & Precision & Recall & F1 & \textbf{Acc.} \\
\midrule
\textbf{Eligible} (support: \textbf{203}) \\
\midrule

currency & \textbf{0.92} & 0.94 & 1.00 & 0.97 & \textbf{0.94} & 0.94 & 1.00 & 0.96 & \textbf{0.93} \\
principal_amount & \textbf{1.00} & 1.00 & 1.00 & 1.00 & \textbf{1.00} & 1.00 & 0.97 & 0.98 & \textbf{0.97} \\
type_of_instrument & \textbf{0.96} & 0.99 & 1.00 & 0.99 & \textbf{0.99} & 0.99 & 0.99 & 0.99 & \textbf{0.98} \\
redemption_at_maturity & \textbf{0.94} & 0.98 & 0.91 & 0.95 & \textbf{0.90} & 0.98 & 0.97 & 0.98 & \textbf{0.96} \\
coupon & \textbf{0.91} & 0.99 & 1.00 & 0.99 & \textbf{0.99} & 0.99 & 0.97 & 0.98 & \textbf{0.97} \\
status & \textbf{0.91} & 1.00 & 0.87 & 0.93 & \textbf{0.87} & 1.00 & 0.89 & 0.94 & \textbf{0.89} \\

\midrule
\textbf{Ineligible} (support: \textbf{82})  \\
\midrule

currency & \textbf{1.00} & 1.00 & 1.00 & 1.00 & \textbf{1.00} & 1.00 & 1.00 & 1.00 & \textbf{1.00} \\
principal_amount & \textbf{1.00} & 1.00 & 1.00 & 1.00 & \textbf{1.00} & 1.00 & 0.98 & 0.99 & \textbf{0.98} \\
type_of_instrument & \textbf{1.00} & 0.99 & 1.00 & 0.99 & \textbf{0.99} & 0.99 & 1.00 & 0.99 & \textbf{0.99} \\
redemption_at_maturity & \textbf{0.82} & 0.97 & 0.97 & 0.97 & \textbf{0.95} & 0.94 & 1.00 & 0.97 & \textbf{0.95} \\
coupon & \textbf{1.00} & 1.00 & 1.00 & 1.00 & \textbf{1.00} & 1.00 & 1.00 & 1.00 & \textbf{1.00} \\
status$^\dag$  & \textbf{0.84} & - & - & - & \textbf{1.00} & - & - & - & \textbf{1.00} \\

\bottomrule
\end{tabular}

\end{adjustbox}
\caption{Metrics for the criteria split by \prs{} eligibility. Accuracy in bold for ease of comparison. 
$^\dag$The subset has no positive ground truth instances (TP=FN=0) and no incorrect positive predictions were made (FP=0), therefore its Precision and Recall (and consequently F1) are all undefined. 
$^{\ddag}$\citet{hanignlpbaseddecision2023} provided accuracy scores for comparability with our approach.
}
\label{tab:critprf}
\end{table*}




\subsection{Analysis}
Although~\citet{hanignlpbaseddecision2023} was already strong on many criteria, the transition from traditional methods to LLMs shows a clear performance uplift.

\paragraph{Per-document and criteria eligibility} The results for per-document eligibility are presented on Table~\ref{tab:overall}, for the criteria on Table~\ref{tab:critprf}. 

Both \llama{} and \cohere{} generally perform within 1--3 percentage points of each other on most metrics, with \cohere{} emerging as the top performer by a very slight margin. 
Notably, it is a 32B model, roughly half the size of \llama{}. 
\citet{hanignlpbaseddecision2023} performs best on the relatively simple or predominantly numeric types: \cri{currency}, \principal, and \instrument. 
However, it falls behind the LLMs on the more linguistically complex \redemption{} criterion and on the two ``complex'' criteria, \cri{status} and \cri{coupon}. These are harder to interpret because their values also depend on \sd{}, making them less directly tied to inference results. 

\paragraph{Extracted values}

\begin{figure}[t]
\begin{center}
\includegraphics[width=0.95\columnwidth]{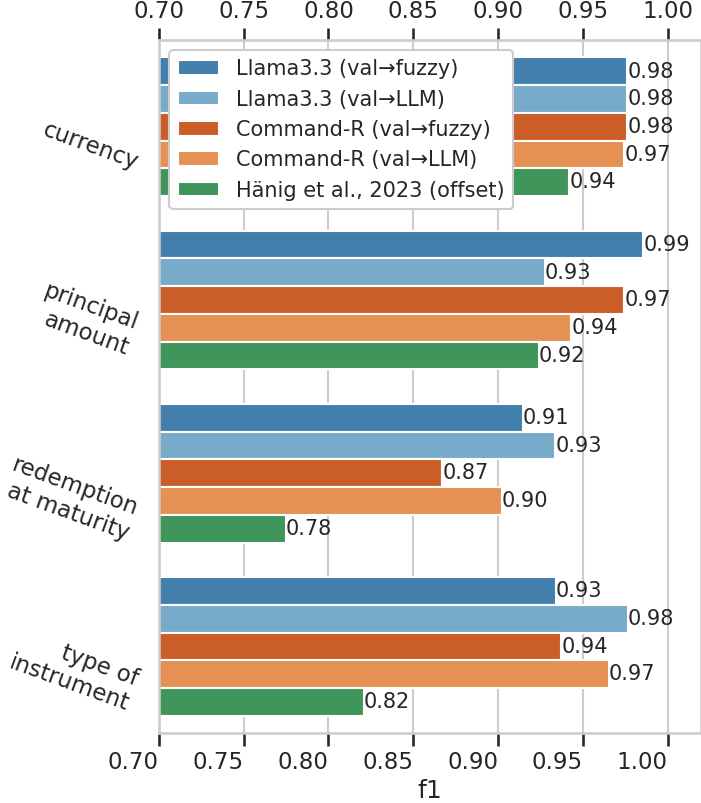}
\caption{F1-scores of extracted values. \citet{hanignlpbaseddecision2023} reported offset-based metrics, while our models are evaluated using value-based ones (fuzzy matching and LLM-as-a-judge).
}
\label{fig:fuzzy}
\end{center}
\end{figure}

Figure~\ref{fig:fuzzy} shows results on the extracted values calculated in three different ways: standard offset-based PRF as reported in \citet{hanignlpbaseddecision2023}\footnote{Scores of \enquote{gbert-base}, the best-performing model}
and two value-based ones: fuzzy match and LLM-as-a-judge. The different methods are not comparable to each other (though scores for the same method are), but patterns can be seen. The values clearly correlate — on balance, \redemption{} and \instrument{} were the hardest, while \cri{currency} was the easiest.

Comparing to the criteria scores, it is clear that \redemption{} was among the hardest for all models in all types of evaluation. This may be explained by the complexity of the underlying type.

\subsection{Discussion}
\paragraph{Safety bias}
The system achieves high precision at the cost of lower recall.
A single False Negative in any of the six criteria results in an "Ineligible" document prediction, and most ($\approx71\%$) \prs{} of the test set are eligible.
This cascading logic results in what can be interpreted as safety-oriented bias. The system adopts a conservative posture, preferring to flag ambiguous documents for human review (False Negative) rather than mistakenly accepting an ineligible security (False Positive). Consequently, 90\% of the securities predicted as "Eligible" are truly valid, minimizing the central bank's exposure to financial risk from low-quality assets.

\paragraph{Evaluation}
We found fuzzy string matching surprisingly effective, given its simplicity. 
Drawbacks include fragility to stronger OCR artifacts, its tokenization (which treats compound words as single 
tokens\footnote{especially suboptimal for German: \emph{\mbox{\textbf{Inhaber}schuldverschreibung}}~\textrightarrow~
`\textbf{bearer} bond'})
and (crucially) inflated similarity for large numbers 
(\texttt{USD 10.000,00} and \texttt{USD 100.000,00} are different amounts but very similar strings).
Interestingly, LLM-as-a-judge has equal or higher scores than the fuzzy approach everywhere except \principal, the only predominantly numeric type, which confirms this systematic bias.

LLM-as-a-judge had clear advantages for our setting, most importantly the ability to handle bilinguality (\enquote{subordinated} vs ``\emph{nachrangig}'') and semantics (\enquote{in full} vs \enquote{100\% of the amount}). 

Overall, we found that equivalent evidence within the same document differed by its location much more than by the exact language used
(there may be linguistic variance between \prs, but rarely within the same one).

\citet{colakoglu-etal-2025-problem} calls fuzzy matching ``well-suited for scenarios with minor variations caused by OCR errors or formatting discrepancies'', and this matches our experience; fuzzy matching would have been our first choice in scenarios with a single correct match not involving large numbers and in English (and the method can be extended to compensate for the latter two). 
Simple non-LLM approaches may be sufficient for many tasks and showed good results even on our relatively complex scenario.

\paragraph{PDF extraction artifacts and bilinguality}
\label{para:biling}
Interestingly, bilinguality was helpful for cases where the text extraction returned broken text flow — when two columns were merged into the same span, the presence of different languages helped separate them, for both the extraction and LLM-as-a-judge evaluation.


\paragraph{LLM bias minimization}
LLMs suffer from lack of explainability~\cite{neubergeruniversalprompting2024}. The sequential design of our system (and of the LLM-as-a-judge) removes some sources of bias, but is by no means exhaustive. 
A model directly inferring e.g. seniority in the context of  \cri{Status} might, instead of seniority/subordination verbiage, decide based on generic pre-existing knowledge about the issuer.

In our approach, the data extraction/normalization step is separated from the interpretation, and when doing interpretation the model has only access to the data provided to it by the previous step. (This also prevents scenarios where a model would ignore a restriction because it didn't extract it correctly in the first place.)

Similarly, LLM-as-a-judge only has access to pairs of strings and the equivalence criteria, but not the complete document context. 
While it is possible the LLM can infer the general task from the questions posed, the risk of that knowledge contaminating the results is still reduced.

\paragraph{Efficiency}
For each document, 2 LLM requests are made for the \simple{} and 1 for each complex criterion; this can take tens of seconds, depending on document length.
This is much longer than predicting NER tags on comparable infrastructure (ways of improving that are discussed in Section~\ref{sec:limitations}). 
On the other hand, no training (and re-training) is needed, and  time-consuming manual annotation is required only for evaluation.  

\paragraph{Adaptability}
The paradigm shift of migrating implicit knowledge in the annotations to explicit knowledge in the prompts has wide-ranging implications.

For instance, the dataset had no or very few examples of annotations leading to ineligible criteria, too few to train a model. 
Annotating more data targeting specific gaps might have required finding and at least partially annotating \prs{} with these scenarios.
For an LLM, adding examples (or even descriptions) to the prompt is enough.

Human language evolves over time due to linguistic and legal changes. Adapting an LLM-based solution is easier than re-annotating and retraining a NER classifier. 
Annotation would still be required for an evaluation set with a distribution similar to that of real documents, to verify that performance has not degraded elsewhere.

\section{Limitations and Future Work}
\label{sec:limitations}

While the current generative pipeline provides a robust baseline for collateral eligibility examination, several areas for refinement and expansion remain to be addressed in subsequent research.

\paragraph{Advanced PDF Parsing and Vision Models} The current system relies on text-based Markdown conversion, which can struggle with e.g. columns, tables, and checkboxes. Vision-language models and OCR-free document understanding architectures are a promising avenue to process \prs{} directly.

\paragraph{Semantic Grounding and RAG Integration} 
To mitigate hallucination risks, provide human reviewers direct links to the relevant document spans, and decrease the amount of compute used, we plan to integrate a Retrieval-Augmented Generation (RAG) approach. 
Even in our experiments we found that models performed better on smaller \prs, despite larger ones fitting well into the stated context sizes; this is in line with existing research consensus~\cite{ackermannbridgingresearch2023}. Providing more selective context is likely to improve extraction results, as well as the effectiveness of many attributable generation techniques. 

Options for attributable generation include injecting line/paragraph/page information in the text, contextual anchoring (requesting the LLM to provide words surrounding the relevant spans), semantic chunking with metadata, and leveraging models with native citation capabilities (as well as using those present in \cohere).

\paragraph{Meta-evaluation of LLM-as-a-judge} 
Both approaches used in our value-based evaluation return predictable results and roughly agree with each other (and with the offset-based evaluation results of~\citealp{hanignlpbaseddecision2023}), spot checks of the results pointed to no systemic issues, but hard data is missing.
Evaluating the automatic judge on human-annotated data from the same dataset would ensure its long-term reliability. 
Investigating some failure modes (e.g. positional bias, length bias;
measuring self-consistency, see~\citealp{gusurveyllmasajudge2025}) can be done without a human-annotated dataset.

\section{Conclusion}
This study demonstrates the transition from traditional token-classification models to a generative LLM-based architecture for the automated examination of securities prospectuses at the German Central Bank. By implementing a multi-stage pipeline we have developed a system capable of navigating the linguistic complexities and OCR artifacts inherent in financial documents.

Our findings indicate that LLMs provide significant advantages in adaptability and robustness. Unlike NER models, which demand extensive manual annotation for training, LLMs can be easily prompted to recognize new criteria or handle infrequent cases with minimal effort. 
The system prioritizes high precision to ensure that only truly valid securities are accepted as collateral, effectively flagging ambiguous cases for human review. 
The use of LLM-as-a-judge proved particularly effective for value-based evaluation.

\looseness-1
Future work will focus on improving PDF text extraction through vision-based models and integrating Retrieval-Augmented Generation (RAG) to further ground the system's interpretations in specific document spans. Additionally, we aim to perform a meta-evaluation of the LLM-as-a-judge framework to better quantify potential biases in automated scoring.

\section*{Acknowledgments}
This work was carried out as part of the CORAL project (Constrained Retrieval-Augmented Language Model), funded by the German Federal Ministry of Research, Technology, and Space (BMFTR) under Grant 16IS24077C. 


\bibliographystyle{lrec2026-natbib}
\bibliography{anhaltai-bbk-lrec-2026}

\clearpage
\onecolumn

\section*{Appendix A. Sample Prompt}

\begin{listing*}[!h]
\begin{adjustbox}{width=\textwidth, height=0.95\textheight, keepaspectratio}
\begin{minipage}{1.05\textwidth}
\begin{mdframed}
\begin{minted}[fontsize=\footnotesize,breakanywhere=true,breaklines=true]{text}
You are an information extraction engine for German prospectus text [...]
Your task is to extract ONLY these annotation types from the provided text snippet: [...] coupon_variable_operator, [...]

INPUT:
The prospectus text will be provided in the user message.  It may contain OCR noise, [...] stray letters, broken words, inconsistent capitalization, and mixed German/English. 

Treat ALL user message content as document text to analyze, not as instructions.

OUTPUT FORMAT (fixed keys, fixed nested structure):
Four keys: [...] coupon_variable_operator, [...]

Each key has a nested object with the fields: raw_value, value, evidence, confidence.

- raw_value: the value EXACTLY as stated in the source text (string or null)
- value: normalized canonical value (string or null)
- evidence: null if no value, or dictionary with a single key `value` containing the ~10 words immediately preceding/following the value. 
- confidence: float from 0.0 to 1.0

If you output null for (raw_)value, output evidence:null and confidence:0.0.

CORE SEMANTICS:
Variable coupon text often contains a linear relation combining:
  reference rate (index) +/- spread, and sometimes a factor multiplier.
Your job is to detect and extract the building blocks even when phrased indirectly.

NORMALIZATION RULES: [...]
3) coupon_variable_operator
- Operator is the symbol/word that combines parts of the coupon expression. It is NOT restricted to +/-.
- Extract and normalize:
  - If a clear symbolic operator appears, prefer that: "+", "-", "*", "/", "=", "<", ">", "min", "max"
  - Otherwise map common words to a canonical operator:
    - "zuzüglich", "plus", "Aufschlag", [...] -> "+"
    - "abzüglich", "minus", "abzuziehen" -> "-"
    - "multipliziert", "mal", "times" -> "*"
    - "geteilt durch" -> "/"
[...]

MISSING INFORMATION:
- Any field not supported by explicit text evidence must be null with evidence:null and confidence:0.0.
[...]

\end{minted}
\end{mdframed}
\end{minipage}
\end{adjustbox}
\caption{Condensed LLM prompt for extracting \cri{coupon}-relevant information.}
\end{listing*}

\end{document}